\documentclass[runningheads]{llncs}

\usepackage{amsmath,amssymb}
\usepackage{graphicx}
\usepackage{booktabs}
\usepackage{url}
\usepackage{listings}
\usepackage{makecell}
\usepackage[caption=false]{subfig}
\usepackage{cleveref}
\crefname{figure}{Fig.}{Figs.}   \Crefname{figure}{Fig.}{Figs.}
\crefname{section}{Sect.}{Sects.} \Crefname{section}{Sect.}{Sects.}
\crefname{equation}{Eq.}{Eqs.}   \Crefname{equation}{Eq.}{Eqs.}
\lstdefinelanguage{sparql}{
  keywords={SELECT,WHERE,PROB,GIVEN,AS,FILTER,INSERT,DELETE,DATA},
  sensitive=true, comment=[l]{\#}, morestring=[b]"
}
\begin{document}

\title{Semantic Bayesian World Models}

\author{Tommaso Soru\orcidID{0000-0002-1276-2366}}
\institute{Liber AI Research, London, UK\\\email{tom@liberai.org}}
\maketitle

\begin{abstract}
Knowledge graphs describe reality in crisp assertions, while the systems now consuming them, foundation models and autonomous agents, reason natively in probabilities.
We argue that this mismatch is why the integration of language models and knowledge graphs remains a data-feeding pipeline rather than a unified reasoning architecture.
We envision Semantic Bayesian World Models (SBWMs): a Web that describes the world not as a database of facts but as a shared, evolving fabric of beliefs over knowledge graphs, where ontological axioms constrain priors, observations update beliefs by Bayesian conditioning, and actions intervene upon the world.
We work through what an agent gains from such a model: a home-security agent deciding whether the figure at the gate is a courier or a burglar, an actuarial estimate aggregated by entailment rather than by string frequency, a planning task that language models reliably fail, and the estimation of quantities that no document has ever stated.
We then set out what the community must build to make them possible: belief annotation over RDF~1.2, probabilistic entailment regimes, semantic calibration layers, and protocols by which agents that have never met can exchange, and disagree over, calibrated beliefs.

\end{abstract}

\section{Introduction}\label{sec:intro}

An autonomous agent must act before it knows. It must decide whether
the figure at the gate is a courier or a burglar, whether a plan will
work, whether a risk is worth taking -- on evidence that is partial,
noisy, and time-critical. Acting under uncertainty requires a
representation \emph{of} uncertainty, and that is precisely what the
Web does not offer the machines now consuming it.

Knowledge graphs describe reality in crisp assertions. A triple is
stated or it is not; an axiom holds or it does not. The open-world
assumption lets a graph model \emph{ignorance} -- what it does not
say -- but not \emph{degrees of belief} -- how strongly we hold what it
does say. That design bought a great deal: shared vocabularies, global
identifiers, machine-checkable entailment, and scale from laboratory
ontologies to graphs of billions of statements. What it cannot express
is the epistemic state of its own consumers.

Those consumers reason natively in probabilities: every token is
sampled from a distribution, and every belief is graded. Yet when they
meet knowledge graphs, the integration is a data-feeding
pipeline -- retrieve triples, paste them into a context window -- rather
than a unified reasoning architecture. We argue that the deeper
obstacle is representational rather than technical: foundation models
traffic in probability distributions, whereas knowledge graphs encode
Boolean assertions, leaving neither able to express the other's
epistemic state natively. The mismatch is measurable: language models
produce probability judgements that violate the axioms of
probability~\cite{zhu2024incoherent} and fall far short of normative
Bayesian belief updating~\cite{qiu2026bayesteach}.

We envision \emph{Semantic Bayesian World Models} (SBWMs): a Web that
describes the world not as a database of facts but as a shared,
evolving fabric of beliefs over knowledge graphs. In an SBWM,
ontological axioms constrain priors, observations update beliefs by
Bayesian conditioning, and actions intervene upon the world. The Web
becomes what Quine and Ullian called a \emph{web of
belief}~\cite{quine1970web} -- except dereferenceable, exchangeable, and
machine-actionable.

We defend, further, a deliberately strong conjecture: language models
cannot scale to superhuman intelligence while knowledge remains
organised as statistical association between substrings. Whatever the
substrate, knowledge must be organised \emph{semantically} and
\emph{probabilistically} -- as propositions with stable identity,
carrying coherent degrees of belief.

The paper proceeds in three movements: what an SBWM is, how it is
represented, and how one could be built (\Cref{sec:sbwm}); what an
agent can do with one that it cannot do with a knowledge graph or a
language model alone (\Cref{sec:atwork}); and what the community must
build to get there (\Cref{sec:agenda}).

\section{Semantic Bayesian World Models}\label{sec:sbwm}

\begin{figure}[t]
  \centering
  \subfloat[\label{fig:venn}]{%
    \includegraphics[width=.48\linewidth]{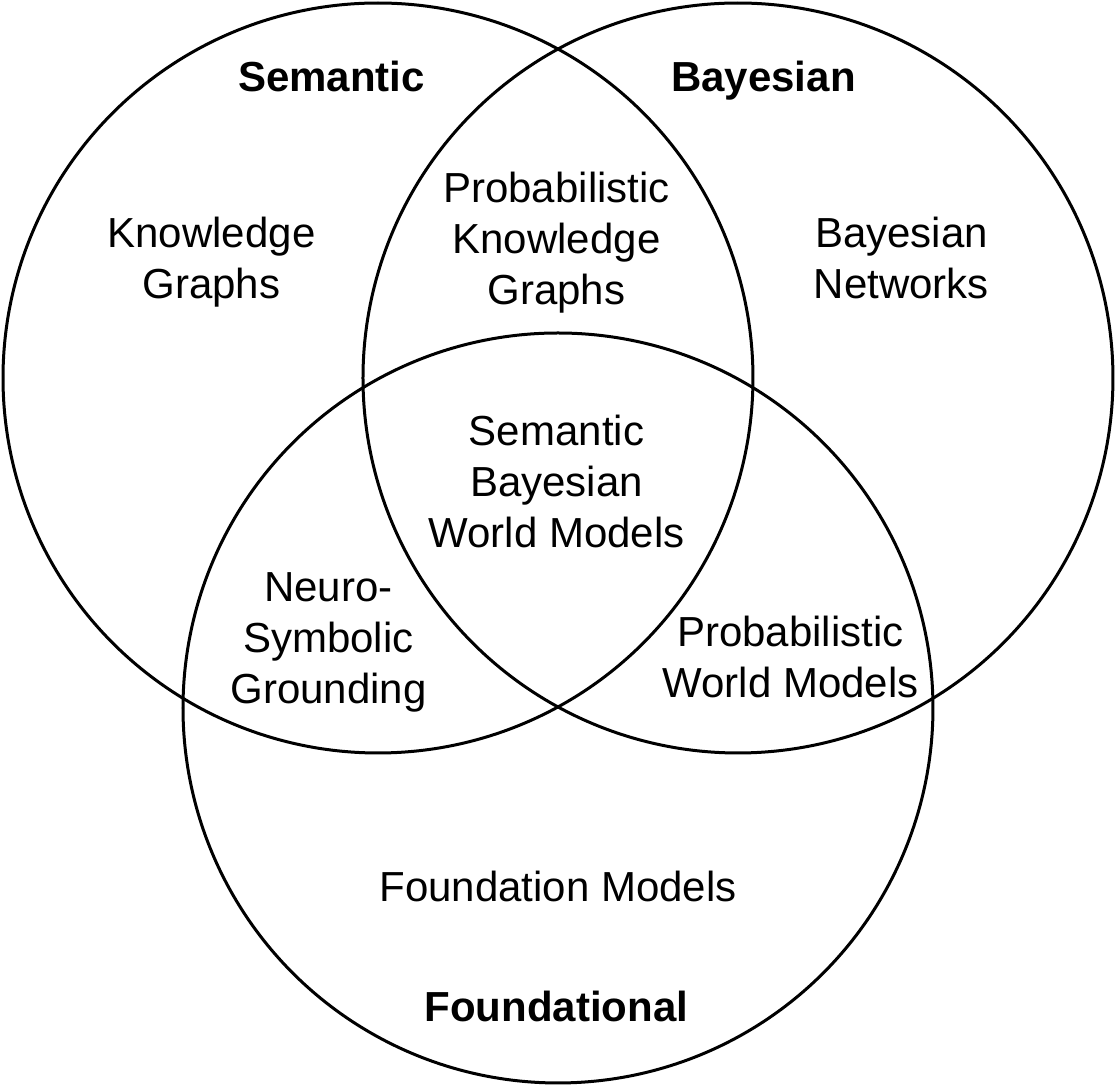}}\hfill
  \subfloat[\label{fig:prior-tensor}]{%
    \includegraphics[width=.48\linewidth]{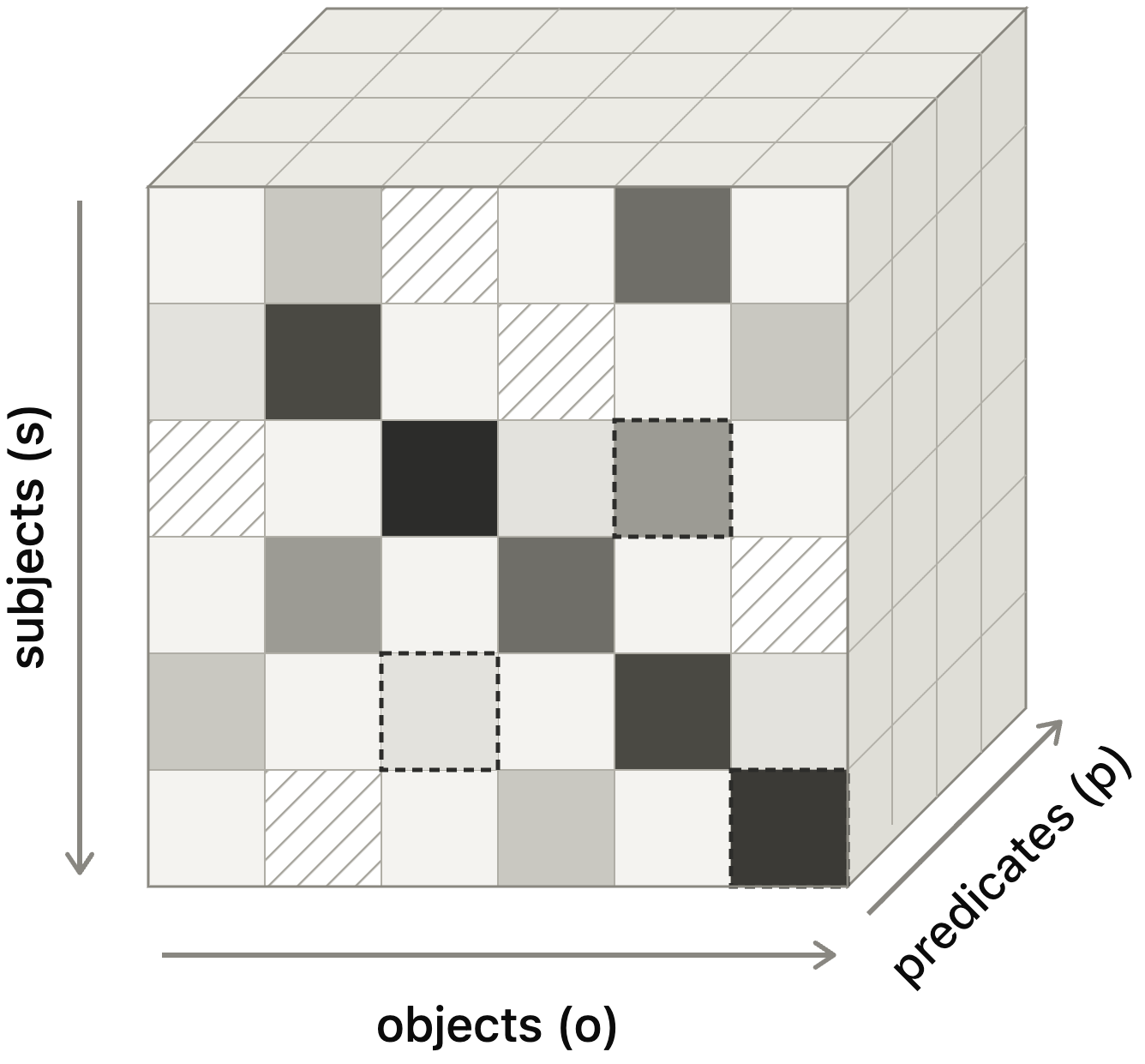}}
  \caption{Two views of a Semantic Bayesian World Model: (a) its
  position at the intersection of three research traditions, and (b) a
  possible computational representation of its graph prior. Cells with diagonal hatch have missing values, some of which are predicted (dashed border) via neural inference.}
  \label{fig:sbwm-overview}
\end{figure}

Three research traditions each hold a piece of the puzzle
(Fig.~\ref{fig:venn}, Table~\ref{tab:capabilities}). \emph{Bayesian
networks} offer calibrated uncertainty, conditional independence, and a
calculus of intervention~\cite{pearl2009causality} -- but their structure
is hand-crafted, their vocabularies are local to each model, and they do
not learn from unstructured data. \emph{Knowledge graphs} offer
web-scale shared semantics, global identity through URIs, and deductive
entailment -- but their statements are crisp, largely static, and silent
about confidence. \emph{Foundation models} learn from everything and
generalise across domains -- but their symbols are ungrounded, their
beliefs sub-symbolic, and their probability judgements
incoherent~\cite{zhu2024incoherent}.

Each pairwise overlap is an active field: probabilistic knowledge graphs
at the intersection of Bayesian networks and knowledge graphs;
probabilistic world models at the intersection of Bayesian methods and
foundation models~\cite{ha2018worldmodels,hafner2023dreamer,lecun2022path,bruce2024genie,yang2026causalworldmodels};
neuro-symbolic grounding at the intersection of knowledge graphs and
foundation models. The three-way centre -- probabilistic, semantic, and
learned -- remains, to our knowledge, unoccupied. That centre is the
Semantic Bayesian World Model.

\begin{table}[h]
  \centering
  \caption{What each tradition can and cannot do.}
  \label{tab:capabilities}
  \begin{tabular*}{\textwidth}{@{\extracolsep{\fill}}lccc@{}}
    \toprule
    Capability & \makecell{Bayesian\\Networks} & \makecell{Knowledge\\Graphs} & \makecell{Foundation\\Models} \\
    \midrule
    Calibrated uncertainty        & \checkmark & --         & partial \\
    Shared web-scale semantics    & --         & \checkmark & -- \\
    Learning from unstructured data & --       & --         & \checkmark \\
    Causal intervention (\emph{do}) & \checkmark & --       & -- \\
    Deductive entailment          & partial    & \checkmark & unreliable \\
    Global identity (URIs)        & --         & \checkmark & -- \\
    Dynamics and prediction       & partial    & --         & \checkmark \\
    \bottomrule
  \end{tabular*}
\end{table}

Every ingredient has been attempted in isolation. The W3C's Uncertainty
Reasoning incubator group mapped the design space~\cite{urw3xg2008};
BayesOWL compiled OWL taxonomies into Bayesian
networks~\cite{ding2004bayesowl}; PR-OWL and MEBN defined probabilistic
ontologies over first-order Bayesian
fragments~\cite{costa2006prowl,laskey2008mebn}; DISPONTE gave
description logics a distribution semantics~\cite{riguzzi2015disponte};
Markov logic, ProbLog, and probabilistic soft logic unified logic with
probabilistic graphical
models~\cite{richardson2006mln,deraedt2007problog,bach2017psl}; noisy
sensors and effectors were given a semantics in the situation
calculus~\cite{bacchus1999noisy}; and a substantial literature managed
uncertainty and vagueness in description
logics~\cite{lukasiewicz2008uncertainty}. What none of these had was a
source of numbers at scale, statement-level annotation as a first-class
citizen, or an application that made graded belief over shared symbols
worth its cost. All three now exist: language-model log-probabilities
beat chance by double digits on real-world
forecasting~\cite{soru2025logprobs}; RDF~1.2, building on the
foundations of RDF-star~\cite{hartig2017rdfstar}, makes statement-level
annotation native; and world models are the
acknowledged frontier of
AI~\cite{ha2018worldmodels,lecun2022path,yang2026causalworldmodels},
which autonomous agents need in order to act.

\subsection{Beliefs Are Predictions}\label{sec:beliefs}

Cognition does not operate on Boolean truth. Even ``this is my car'' is
a belief held with very high confidence, not a theorem. Agents survive
by predicting their environment and minimising prediction
error -- a view developed most ambitiously in the free-energy
principle~\cite{friston2010fep}, whose neuroscientific claims remain
contested but whose engineering slogan we happily borrow: \emph{a belief
is a prediction, and learning is belief revision}.

We define an SBWM as a tuple $\mathcal{M} = (\Sigma, P_0, T, O)$:
a vocabulary and TBox $\Sigma$; a prior $P_0$ over RDF graphs; a
transition kernel $T(G' \mid G, a)$ over graph edits, capturing how
actions $a$ change the world; and an observation model $O(o \mid G)$
mapping noisy perception -- sensor readings, extractor outputs, language
model log-probabilities -- to likelihoods over triples. The belief state
is a distribution over graphs; Bayes' rule is the update; the ontology
is the prior.

\subsection{Ontologies Are Priors}\label{sec:axioms}

The organising principle is \emph{coherence with respect to entailment}:
\begin{equation}
  A \models B \;\Longrightarrow\; P(A) \le P(B),
\end{equation}
from which the TBox yields a family of constraints for free:
\begin{align}
  \langle c_2, \mathtt{rdfs{:}subClassOf}, c_1\rangle
    &\;\Longrightarrow\; P(\langle s, \mathtt{rdf{:}type}, c_2\rangle)
       \le P(\langle s, \mathtt{rdf{:}type}, c_1\rangle) \label{eq:subclass}\\
  \langle p_2, \mathtt{rdfs{:}subPropertyOf}, p_1\rangle
    &\;\Longrightarrow\; P(\langle s, p_2, o\rangle)
       \le P(\langle s, p_1, o\rangle) \label{eq:subprop}\\
  \langle c_1, \mathtt{owl{:}disjointWith}, c_2\rangle
    &\;\Longrightarrow\; P(\langle s, \mathtt{rdf{:}type}, c_1\rangle
       \wedge \langle s, \mathtt{rdf{:}type}, c_2\rangle) = 0 \\
  \mathrm{dom}(p) = c
    &\;\Longrightarrow\; P(\langle s, \mathtt{rdf{:}type}, c\rangle
       \mid \langle s, p, o\rangle) = 1 \label{eq:domain}
\end{align}
These constraints are more useful than they look, because refinement
can only subtract mass:
\begin{equation*}
  P(\textit{man}) \;\ge\; P(\textit{man} \wedge \textit{called Andrea})
  \;\ge\; P(\textit{man} \wedge \textit{called Andrea}
             \wedge \textit{likes Ferrari}),
\end{equation*}
a bound any estimator must respect. Eqs.~\ref{eq:subclass}
and~\ref{eq:subprop} impose the same shape inside the graph:
\texttt{:Man rdfs:subClassOf :HumanBeing} caps the belief that $x$ is a
man by the belief that $x$ is a human being, and
\texttt{:favouriteCar rdfs:subPropertyOf :likes} caps the belief that a
Ferrari is $x$'s favourite car by the belief that $x$ likes Ferraris.
The bounds transfer across languages because they are stated over URIs:
that
\emph{Andrea} names a man in Italian and a woman almost everywhere else
is a fact no credence attached to a string survives, and one a credence
attached to an identifier never meets.

Two consequences follow. First, hierarchical priors defined along the
class tree satisfy Eq.~\ref{eq:subclass} \emph{by construction} and give
unseen subclasses inherited prior mass from their parents: the ontology
computes priors for data never observed. Second, any neural scorer can
be made coherent by projecting its outputs onto the polytope defined by
the axioms -- if perception reports $P(\mathit{cup})=0.9$ but
$P(\mathit{container})=0.6$, the projection repairs the violation. Such
a \emph{semantic calibration layer} is differentiable and drops into any
architecture; de Finetti's coherence argument supplies its normative
justification~\cite{definetti1974}.

\subsection{Adapting SPARQL to Causal Inference}\label{sec:causality}

The distinction at the heart of causal inference -- seeing versus
doing~\cite{pearl2009causality} -- maps directly onto the existing stack:
SPARQL \texttt{WHERE} is conditioning; SPARQL \texttt{UPDATE} is the causal
\emph{do}-operator, with \texttt{DELETE}/\texttt{INSERT} performing
graph `surgery' on the world graph itself. RDF~1.2 carries the belief
annotations on the statements themselves,
\begin{center}
\texttt{<<\,:x001 :likes :Ferrari\,>> :prob "0.3"\^{}\^{}xsd:decimal .}
\end{center}
and conditional structure lives in lifted networks whose nodes are
triple patterns, in the tradition of MEBN
fragments~\cite{laskey2008mebn}, with conditional probability tables
published as Linked Data. The ontology contributes conditionals for free
(Eq.~\ref{eq:domain}). An agent-native query surface then falls out
naturally:

\begin{lstlisting}[language=sparql,caption={A conditional-probability
query over an SBWM.},label={lst:prob}]
SELECT ?c (PROB { ?x rdf:type ?c }
           GIVEN { ?x :locatedIn :Kitchen } AS ?p)
WHERE { ?c rdfs:subClassOf :Container }
\end{lstlisting}

\subsection{Representation, Learning, and Scale}\label{sec:representation}

A distribution over graphs admits a concrete representation: sparse
tensors indexed by subject, predicate, and object. Priors
$P(\langle s,p,o\rangle)$ form a 3-dimensional tensor (\Cref{fig:prior-tensor}); one-event
conditionals $P(\langle s,p,o\rangle \mid \langle s_1,p_1,o_1\rangle)$ a
6-dimensional one; two-event conditionals a 9-dimensional one; and so
on. These tensors are astronomically sparse, and therein lies a potential new
research avenue: \emph{semantic Bayesian tensor completion}. Missing
entries are predicted from semantically adjacent ones -- knowledge graph
completion graduates from link prediction to prior
estimation~\cite{ren2020betae} -- and efficient Bayesian update becomes a
question of sparse tensor algebra.

The gain is the ability to answer questions no source has ever stated.
Suppose an agent needs
$P(\langle x, \mathtt{{:}hasVisited}, \mathtt{{:}Ibiza}\rangle \mid
\mathtt{age}(x)\!\in\![20,22])$. No document reports it and no cell
holds it, so a language model can only interpolate between phrasings it
has seen. In a tensor organised by concepts, the neighbours are
identifiable: the adjacent bracket is known,
$P(\cdot \mid [18,20]) = 0.33$, and the age profile of the sibling and
parent classes of \texttt{:Ibiza} under
\texttt{:LeisureDestination} is known from cells that \emph{were}
observed. Completing the entry is a transport problem over semantically
adjacent cells, and what comes back is a posterior with provenance
rather than a fluent sentence.

The substrate already scales:
trillion-triple loads have been demonstrated repeatedly, by AllegroGraph
in 2011,\footnote{\url{https://www.w3.org/wiki/LargeTripleStores}.} by Oracle at 1.08
trillion edges~\cite{oracle2021trillion}, and by Stardog across
clouds,\footnote{\url{https://www.stardog.com/blog/trillion-edge-knowledge-graph/}.}
on resources modest beside those used to serve a single frontier
language model.

\subsection{Building One}\label{sec:building}

Construction needs no component that does not already exist.
Multilingual web text is processed by small language models fine-tuned
for knowledge extraction, or by dependency parsers, into typed triples;
the extractor's own log-probabilities, rather than a number elicited by
hand, supply the initial confidence on each
statement~\cite{soru2025logprobs,soru2024trend}, and translating
language into probabilistic programs is an increasingly practical
route~\cite{wong2023word}. Aggregating across documents and
languages -- where the same proposition extracted from a German and a
Portuguese source is one URI, not two strings -- yields a \emph{semantic
Bayesian knowledge graph}: a graph whose every statement carries a
credence with provenance. Semantic Bayesian tensor completion then
estimates the cells no document supports, turning the graph into a
usable prior $P_0$; adding a transition kernel and an observation model
turns that prior into a world model. Each stage is a recognisable
research task. What is missing is the commitment to carry the
probabilities through all of them, instead of thresholding them away at
the first step -- which is what today's extraction pipelines do, and why
the confidences they compute never reach the agent that needs them.

\section{SBWMs at Work}\label{sec:atwork}

\subsection{A Camera in the Garden}\label{sec:camera}

A household security agent runs locally on a camera overlooking a
garden. At 22:04 it detects a person at the gate, carrying a box.
Raise the alarm, or not?

Perception alone cannot answer, because the question is about an
unobservable: the visitor's goal. The vision model reports a person
($0.99$), a box-shaped object ($0.82$), and no uniform clearly visible
($0.4$ that one is present); nothing in that vector distinguishes a
late delivery from a burglary. A language model asked the question in
prose will produce a fluent answer attached to a number that moves when
the prompt is paraphrased.

An SBWM answers by decomposition. The unobservable in question is the
visitor's goal, $\langle ?p, \mathtt{{:}goal}, \mathtt{{:}Theft}\rangle$,
and Bayes' rule turns one unanswerable question into several answerable
ones,
\begin{equation}
  P(\mathit{Theft} \mid o) \;=\;
  \frac{P(o \mid \mathit{Theft})\,P(\mathit{Theft})}
       {\sum_{h \in \mathcal{H}} P(o \mid h)\,P(h)},
  \qquad
  \mathcal{H} = \{\mathit{Theft}, \mathit{Delivery}, \mathit{Visit},
  \dots\},
\end{equation}
each of which is a belief over a triple with a stable identifier, and
each of which comes from a different publisher
(Table~\ref{tab:camera}).

\begin{table}[h]
  \centering
  \caption{Beliefs the agent needs, and where each one comes from.
  No two share a publisher; all share a vocabulary.}
  \label{tab:camera}
  \begin{tabular*}{\textwidth}{@{\extracolsep{\fill}}p{5.4cm}p{6.0cm}@{}}
    \toprule
    Belief & Source \\
    \midrule
    $P(\langle ?p, \mathtt{{:}goal}, \mathtt{{:}Theft}\rangle)$
      & burglary rate for the postcode, from a police open-data
        endpoint, conditioned on month and hour \\
    $P(\langle \mathtt{{:}order42}, \mathtt{{:}deliveryDue}, \mathtt{today}\rangle)$
      & the household's own graph \\
    $P(\langle ?p, \mathtt{rdf{:}type}, \mathtt{{:}Courier}\rangle \mid \mathtt{hour}=22)$
      & the carrier's published delivery-hour distribution \\
    $P(\langle \mathtt{{:}neighbour1}, \mathtt{{:}isAt}, \mathtt{{:}Home}\rangle)$
      & presence signals next door, shared under an access policy \\
    $P(o \mid \langle ?x, \mathtt{rdf{:}type}, \mathtt{{:}Parcel}\rangle)$
      & the camera's own vision model, as $O(o \mid G)$ \\
    \bottomrule
  \end{tabular*}
\end{table}

Three properties of this arrangement are unavailable to a monolithic
model. First, the beliefs are \emph{separately sourced}: the crime rate
comes from the police, the delivery window from the carrier, the order
from the household. None of these parties trained a model together, and
none needs to; they publish beliefs at dereferenceable URIs over a
shared vocabulary, and the agent merges them under explicit provenance.
Second, they are \emph{separately updatable}: when the household
cancels the order, exactly one credence changes and the posterior moves
with it -- no retraining, no prompt engineering, and an audit trail
showing which belief did the work. Third, the ontology keeps the
hypotheses honest: with \texttt{:Courier owl:disjointWith :Burglar} and
both \texttt{rdfs:subClassOf :Visitor}, the competing explanations are
forced to compete for mass, and Eq.~\ref{eq:subclass} guarantees
$P(\mathit{Visitor}) \ge P(\mathit{Courier})$ however the neural scorer
behaves.

The agent can also ask what to \emph{do}. Switching on the floodlight
is an intervention, not an observation, and the distinction is
material: the agent evaluates it by applying
\texttt{INSERT DATA \{ :floodlight :status :On \}} to the world
graph -- Pearl's $do(\cdot)$, realised as graph surgery
(\Cref{sec:causality}) -- and reading off
$P(\langle ?p, \mathtt{{:}leaves}, \mathtt{{:}Property}\rangle)$ under each
hypothesis through the transition kernel $T$. A courier does not flee a
floodlight; a burglar does. The action that best discriminates between
the hypotheses is therefore computed, not prompted, and the alarm is
raised on a posterior the household can inspect.

\subsection{Aggregation by Entailment}\label{sec:aggregate}

An insurer's agent needs the probability that a vehicle of a given
class, driven by a driver in a given age band, is involved in an
accident in the rain. A language model's estimate reflects how often
near-identical sentences occurred in training text: synonyms shift it,
translation shifts it, and instances that must be \emph{inferred} -- a
vehicle typed only by its model name, whose class follows by
entailment -- are missed entirely. If the model is semantic, cases are
aggregated by entailment instead: one graph pattern covers every
subclass and every instance, each make of car and each language of
report. If the model is Bayesian, the estimate is a posterior that
moves as new evidence arrives, except that in an SBWM the likelihoods
attach to concepts rather than to strings, so a claim filed in
Portuguese updates the same belief as a claim filed in German.

\subsection{The Car Wash Test}\label{sec:carwash}

Some failures are about planning rather than estimation. Consider:
``My car needs washing, but the car wash is only 100\,m away. Should I
walk or drive?'' State-of-the-art language models frequently answer
\emph{walk} -- the distance is short -- missing that washing requires the
car to be at the car wash. A few triples of formalisation dissolve the
confusion:

\begin{lstlisting}[caption={The car wash test, formalised.},label={lst:carwash}]
:washes rdfs:domain :CarWash ; rdfs:range :Car ;
        rdfs:subPropertyOf :sharesLocationWith .
:carwash001 a :CarWash .
:car001 a :Car ; :status :Dirty ; :owner :Me ;
        :sharesLocationWith :Me .
# Rule: ?o :status :Clean <- ?s :washes ?o .
# Goal: :car001 :status :Clean .
\end{lstlisting}

Backward chaining from the goal requires some
$\langle ?s, \mathtt{{:}washes}, \mathtt{{:}car001}\rangle$; the domain
axiom forces $?s$ to be a car wash; the sub-property axiom forces
$\langle ?s,$ $\mathtt{{:}sharesLocationWith}, \mathtt{{:}car001}\rangle$;
and since the car currently shares a location with its owner, not with
the car wash, any plan must move the car, i.e. \emph{drive}. Under uncertain
perception the same machinery degrades gracefully: if a vision system
is only $90\%$ confident the object is a car, every downstream belief
reflects that uncertainty instead of relying on an overconfident binary assertion.


\subsection{Why Symbols Matter}\label{sec:riddles}

The three vignettes share a structure. In each, the agent must isolate
a belief, source it, update it, and aggregate it over everything the
belief entails. When knowledge is stored as statistical association
between strings, beliefs cannot be isolated and updated individually,
credences attach to phrasings rather than propositions, and entailment
is approximated by similarity. Superhuman performance at forecasting,
planning, and science demands exactly the operations this
representation denies. Scaling parameters and data sharpens the
approximation without changing the representation; even models
explicitly taught normative updating acquire the skill approximately
and sub-symbolically~\cite{qiu2026bayesteach}.

We therefore argue that
language models cannot scale to superhuman intelligence without
organising knowledge in a semantic and probabilistic framework:
propositions with stable, language-invariant identity, carrying
credences that obey the axioms of probability. The claim is
falsifiable -- a pure language model that maintained coherent credences
under paraphrase and translation, and aggregated them across entailed
instances, would refute it~\cite{zhu2024incoherent}. And the framework
need not be an external triple store: if fragments of one already exist
implicitly inside the weights, the case for building it
explicitly -- auditable, shareable, dereferenceable -- only strengthens.


\section{What Must Be Built}\label{sec:agenda}

The vision demands precise architectural shifts.
First, a W3C \emph{belief-annotation vocabulary} -- PROV for priors -- so
that probabilities, their calibration method, and their provenance
travel together. Second, \emph{probabilistic entailment regimes}, in
which classical entailment is the probability-one special case and
subsumption acts as a monotonicity constraint (Eq.~\ref{eq:subclass}).
Third, \emph{probabilistic SHACL}, reading shapes as soft constraints
with violation costs rather than binary conformance. Fourth,
\emph{semantic calibration layers} as standard components between neural
scorers and triple stores (\Cref{sec:axioms}). Fifth, \emph{federated
belief exchange}: priors published at dereferenceable URIs, merged
under explicit provenance, so that two agents who have never met can
disagree \emph{numerically} -- and resolve the disagreement by evidence.
The camera in the garden needs all five, and needs nothing else.


The agenda is ambitious because its hardest questions
remain open. Identity must itself become probabilistic: uncertain
coreference and \texttt{owl:sameAs} cannot simply be assumed away.
Tractability is equally fundamental. Weighted model counting is
\#P-hard, higher-order conditional tensors grow combinatorially, and
practical systems will depend on sparsity, factorisation, and lifted
inference over ontology symmetries. The numbers also require scrutiny:
logit-derived confidences vary across models and
phrasings~\cite{zhu2024incoherent}, so calibration methods and source
reliability must travel with the beliefs they produce.

These are not peripheral caveats but the criteria by which the vision
should be judged: a probabilistic Web must deliver calibration and
coherence, not merely accuracy. Yet nothing above is distant. The parts
already exist -- extractors that emit confidences, stores that annotate
statements, ontologies that constrain them -- held apart only by the
habit of discarding probabilities at the first opportunity. An agent
deciding whether to raise the alarm cannot afford that habit, and the
Semantic Web is best placed to spare it: once it learns to say not
merely what is the case, but how strongly it is believed, and on what
evidence.

\bibliographystyle{splncs04}
\bibliography{references}

\end{document}